# Extend the shallow part of Single Shot MultiBox Detector via Convolutional Neural Network

Liwen Zheng, Canmiao Fu, Yong Zhao[*]
School of Electronic and Computer Engineering,
Shenzhen Graduate School of Peking University, Shenzhen, China

## ABSTRACT

Single Shot MultiBox Detector (SSD) is one of the fastest algorithms in the current object detection field, which uses fully convolutional neural network to detect all scaled objects in an image. Deconvolutional Single Shot Detector (DSSD) is an approach which introduces more context information by adding the deconvolution module to SSD. And the mean Average Precision (mAP) of DSSD on PASCAL VOC2007 is improved from SSD's 77.5% to 78.6%. Although DSSD obtains higher mAP than SSD by 1.1%, the frames per second (FPS) decreases from 46 to 11.8. In this paper, we propose a single stage end-to-end image detection model called ESSD to overcome this dilemma. Our solution to this problem is to cleverly extend better context information for the shallow layers of the best single stage (e.g. SSD) detectors. Experimental results show that our model can reach 79.4% mAP, which is higher than DSSD and SSD by 0.8 and 1.9 points respectively. Meanwhile, our testing speed is 25 FPS in Titan X GPU which is more than double the original DSSD.

## 1. INTRODUCTION

Object detection holds great significance for computer vision and image processing field because it has a wide range of applications in video surveillance, intelligent transportation, medical diagnosis and military visual guidance. Therefore, to have a robust and fast object detection algorithm will be of great significance. To deal with this task, previous deep learning methods' achievement was heavily driven through the power of the Convolutional Neural Networks (CNN). Of the many algorithms based on CNN, the two branches are particularly noticeable. First one is the R-CNN series, including R-CNN [1], SPPnet [2], Fast R-CNN [3], Faster R-CNN [4], which use two stage to solve the problem. In this series, the first step is used to find possible candidate regions and the second step is to predict the corresponding category and perform bounding box regression on the candidate regions. Second is single stage series, including You Only Look Once (YOLO) [5], YOLOv2 [6], Single Shot MultiBox Detector (SSD) [7], which aim to remove the region proposal stage and then predict confidence and offset for every default box. Taking the two stage method further by introducing more context information for each layer which is used to predict output, the Deconvolutional Single Shot Detector (DSSD) [8] is currently one of the best models in terms of detection accuracy and speed at that time. It seems like that by replacing the VGG [9] network with too much deeper Residual-101 [10] network and add too much extra deconvolution layers to get high-level context,

*yongzhao@pkusz.edu.cn;

DSSD [8] still inevitably loses too much speed in order to improve accuracy. Therefore, for a high-precision detector, increasing the speed without degrading its accuracy, is still a challenging problem.

In this paper, we propose a way to enhance the shallow part of SSD, which can make our model faster and more accurate than DSSD. In order to achieve this, we first use the SSD as our base network, which can achieve 77.5% mAP on PASCAL VOC2007 test and 46 FPS detection speed performance. After that, we apply our extension module to the first three SSD layers. Finally, we improve the prediction modules for these three changed SSD layers. The rest of the paper is organized as follows: Section 2 introduces the method we proposed. Section 3 gives the experimental results and discussion. The last section is the conclusion.

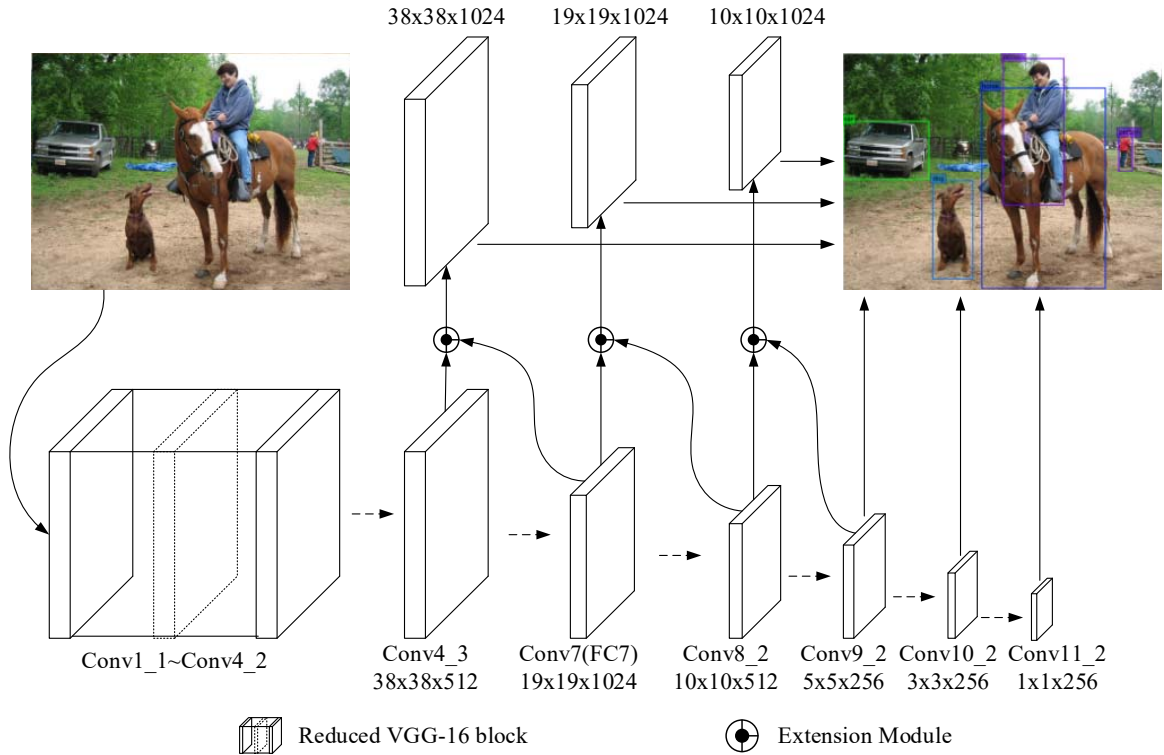

Figure 1. The framework of our ESSD.

## 2. OUR APPROACH

The ESSD model we proposed is illustrated in Figure 1. In this section, we first review the structure of SSD and then introduce our extension module and explain why our approach has such a good effect.

### 2.1 SSD

The Single Shot MultiBox Detector (SSD) [7] uses the reduced VGG-16 [9] as based network and add extra convolution layers to the end of it. Then selecting some layers of different sizes from extra added convolution layers and based layers to predict scores and offsets for some default boxes on each scale. These predictions are generated by a single convolution layer, and the number of convolution kernels for these layers is related to the number of default boxes and the number of prediction categories.

As RRC [11] and RON [12] mentioned, the semantic information of shallow layers of SSD is limited, resulting in the performance of shallow layers in detecting small object is far weaker than the performance of large ones in deep layers. Based on the above, we believe that extending semantic information to shallow part of SSD can improve the accuracy of detection.

**2.2 Extension Module**

Although deconvolution approaches have been very successful in the task of extracting the semantic information of the target being detected [8, 11, 13], adding too much deconvolution layers will inevitably lead to a significant increase in computational time-consuming [8, 11], while providing too little semantic information will certainly lose some detection performance [13]. Based on the above consideration, three extension modules are applied in our proposed framework as shown in Figure 1, which have extended right amount of semantic information and can get good enough results without losing too much detection speed. The extension module is shown in Figure 2.

For our extension module, firstly we use a deconvolution layer to make the Conv layer n+1 the same size as Conv layer n, and a convolution layer followed with batch normalization layer and ReLU layer are applied to refine the features we learned. Then two Conv-bn-relu modules as described above are behind the Conv layer n to get the same learning ability as Conv layer n+1. Finally, we use one extended operation to obtain the extension features, which can be concatenation, elementwise-sum, or elementwise-product. Inspired from DSSD [8], which adds residual blocks to each prediction layer, we only add one extra 1x1x512 convolution layer to each prediction layer in order not to make the model too complicated. Our experiment shows that this extra convolution layer can improve 0.4% mAP.

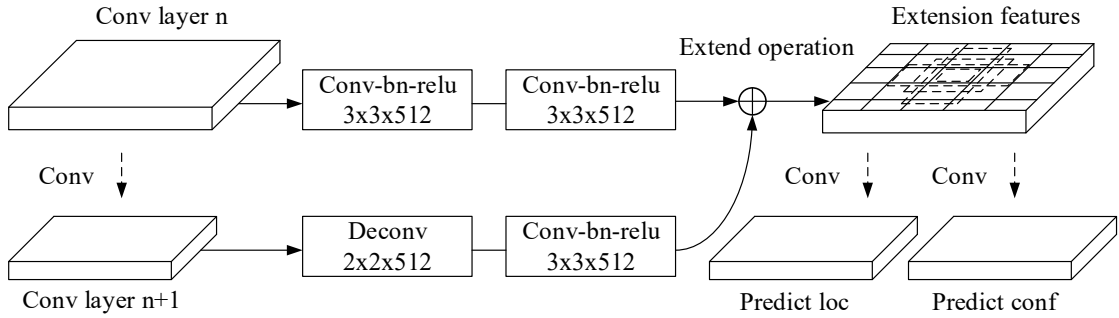

Figure 2. Extension module.

After adding the extension module, Conv4_3, Conv7 (FC7), Conv8_2 and Conv9_2 can receive gradient backpropagation from multiple layers during the training phase, and the features learned in this way will have richer semantic information. Connecting layers in this way makes our model even more advantageous when compared to DSSD.

**2.3 Weighted average depth**

In order to analyze the difference between our ESSD and SSD, in the context that the existing description of network depth is limited, we are inspired by the notion of a weighted average and propose the concept of weighted average depth. The weighted average depth of a layer can be defined as follows,

$$\bar{D}(L_n) = \begin{cases} 1 & \text{if layer n directly connected to data,} \\ \sum_{i=1}^{k} L_n^i w_i & \text{otherwise.} \end{cases} \quad (1)$$

Where $L_n^i$ is the ith input layer of layer $L_n$, $w_i$ is the weights of the ith input layer. In the SSD and ESSD models, only the weight of each input layer in the elementwise operations layer is 0.5, and in other cases the weights are all 1. In this way, we obtain the weighted average depth of each of the layers used to predict conf and loc in SSD and ESSD. The depths are shown in Table 1.

Since we want to get good enough object detection results from these feature layers, the weighted average depth of these feature layers should not be much too different. So we use the coefficient of variation as a comparative indicator of the two models. According to the data in Table 3, the coefficients of variation of SSD and ESSD are 24.19% and 13.72%. This indicates that the weighted average depth between SSD varies greatly, while the differences between ESSD are small. Therefore, every layer of ESSD that used to predict can learn good features, which explains why the ESSD model has such a good effect.

Table 1. Weighted average depth of layers at different scales which used to predict conf and loc.

| Method | 38x38 | 19x19 | 10x10 | 5x5 | 3x3 | 1x1 |
|---|---|---|---|---|---|---|
| SSD | 10 | 15 | 17 | 19 | 21 | 23 |
| ESSD | 14.5 | 18 | 20 | 19 | 21 | 23 |

## 3. EXPERIMENT

Our model is trained on the union of PASCAL VOC2007 trainval and PASCAL VOC2012 trainval that has 20 object categories, and evaluate the results on PASCAL VOC2007 test (4952 images). And it is built on original SSD [7] implementation which is based on the Caffe [14]. We evaluate object detection accuracy by mean Average Precision (mAP).

Table 2. PASCAL VOC2007 test detection results.

| Method | mAP | aero | bike | bird | boat | bottle | bus | car | cat | chair | cow | table | dog | horse | mbike | person | plant | sheep | sofa | train | tv |
|---|---|---|---|---|---|---|---|---|---|---|---|---|---|---|---|---|---|---|---|---|---|
| Faster [4] | 73.2 | 76.5 | 79.0 | 70.9 | 65.5 | 52.1 | 83.1 | 84.7 | 86.4 | 52.0 | 81.9 | 65.7 | 84.8 | 84.6 | 77.5 | 76.7 | 38.8 | 73.6 | 73.9 | 83.0 | 72.6 |
| SSD300 [7] | 77.5 | 79.5 | 83.9 | 76.0 | 69.6 | 50.5 | **87.0** | 85.7 | 88.1 | 60.3 | 81.5 | 77.0 | 86.1 | 87.5 | 84.0 | 79.4 | 52.3 | 77.9 | 79.5 | 87.6 | 76.8 |
| DSSD321 [8] | 78.6 | 81.9 | 84.9 | **80.5** | 68.4 | 53.9 | 85.6 | 86.2 | **88.9** | 61.1 | 83.5 | **78.7** | 86.7 | **88.7** | 86.7 | 79.7 | 51.7 | 78.0 | **80.9** | 87.2 | **79.4** |
| ESSD-less (ours) | 79.0 | 79.7 | **86.6** | 79.1 | **72.5** | 54.8 | 86.7 | **87.0** | 87.9 | 63.0 | 82.9 | 76.2 | 86.9 | 87.9 | **87.4** | 79.8 | 55.6 | 80.0 | 79.5 | 88.0 | 79.3 |
| ESSD-prod (ours) | 78.8 | 82.2 | 83.8 | 78.4 | 72.4 | 53.2 | 86.7 | 86.8 | 87.7 | 62.2 | 84.4 | 75.4 | 87.0 | 87.5 | 86.1 | 79.5 | 54.9 | 81.1 | 80.3 | 88.1 | 78.4 |
| ESSD-concat (ours) | 79.1 | 81.6 | 86.8 | 78.9 | 72.4 | 53.6 | 86.6 | 86.7 | 87.5 | **63.1** | 83.6 | 76.2 | 86.1 | 87.9 | 86.4 | 79.8 | 55.2 | **81.2** | 80.4 | 88.4 | 78.7 |
| **ESSD-sum (ours)** | **79.4** | **82.6** | 86.1 | 79.8 | 72.2 | 54.7 | 86.8 | 86.9 | 88.2 | 62.8 | **85.2** | 78.2 | **87.5** | 88.0 | 87.0 | **80.0** | **56.1** | 80.2 | 80.4 | **88.7** | 78.1 |

### 3.1 PASCAL VOC2007

Our ESSD training process consists of three steps: starting with training the original SSD, then training extra added layers, and finally fine-tuning the entire network. For the first step, we use the batch size of 32, and starts the learning rate at $10^{-3}$ for the first 80K iterations, then continue training for 20K iterations with $10^{-4}$ and 20K iterations with $10^{-5}$. We use this as our pre-trained model. For the second step, we only train the extra added layers to make them well integrated. We use the learning rate at $10^{-3}$ for the first 20K iterations and decrease it to $10^{-4}$ for the next 25K iterations. For the third step, we fine-tune the entire network. We set the learning rate at $10^{-3}$ for the first 20K iterations, then continue training for 20K iterations with $10^{-4}$ and 30K iterations with $10^{-5}$ and 20K iterations with $10^{-6}$.

The results on PASCAL VOC2007 test detection are shown in Table 2. It describes the impact of our additions on improving the detection performance. The original SSD without our extension module can get 77.5% mAP. When adding three of our extension module, the ESSD-less can reach 79.0% mAP. Finally, an extra convolution layer is added to the layers used to predict conf and loc, we obtain the better result of 79.4% mAP, which is higher than DSSD and SSD by 0.8 and 1.9 points respectively. The ESSD-prod and ESSD-concat in Table 2 show that both of elementwise-product and concatenation extend operation cannot get better performance than elementwise-sum.

### 3.2 Inference Time

We measure the speed with batch size 1 using Titan X and cuDNN 6.0.21 with Intel(R) Core(TM) i7-4790K@4.00GHz, as shown in Table 3. Our ESSD maintains a great speed advantage over R-FCN [15] and DSSD513 [8], but the accuracy is only slightly lower. Although our ESSD is slower than the original SSD, it outperforms DSSD321 [8] and Faster R-CNN [4] in both detection speed and accuracy.

Table 3. The running time on PASCAL VOC2007 test.

| Method | mAP | FPS | batch size | Input resolution |
|---|---|---|---|---|
| Faster R-CNN (VGG 16) [4] | 73.2 | 7 | 1 | ~1000×600 |
| R-FCN [15] | 80.5 | 9 | 1 | ~1000×600 |
| YOLOv2 [6] | 76.8 | 67 | 1 | 416×416 |
| YOLOv2 544×544 [6] | 78.6 | 40 | 1 | 544×544 |
| SSD300 [7] | 77.5 | 46 | 1 | 300×300 |
| SSD512 [7] | 79.5 | 19 | 1 | 512×512 |
| DSSD321 [8] | 78.6 | 11.8 | 1 | 321×321 |
| DSSD513 [8] | 81.5 | 5.5 | 1 | 513×513 |
| ESSD (ours) | 79.4 | 25 | 1 | 300×300 |

### 3.3 Visualization

We show some detection results with scores above 0.6 on PASCAL VOC2007 test with SSD300 and ESSD. As we can see from Figure 3, our ESSD has two distinct improvements based on original SSD. The first is the improvement of the detection of small targets, mainly because our extension module provides rich semantic information for the shallow layers that are responsible for detecting small objects. The second is that the ESSD has a higher score for the object which can be detected by both models, mainly because our more sophisticated model have greater learning ability than SSD.

## 4. CONCLUSION

This paper proposes an approach to provide semantic information for the shallow layers of the SSD. Our ESSD300 model outperforms DSSD321 [8] and Faster R-CNN [4] in both detection speed and accuracy which can reach 79.4% mAP on PASCAL VOC2007 test and up to 25 FPS on a single Titan X GPU. Moreover, we propose the concept of weighted average depth to describe the difference between SSD300 and ESSD300 and explain why our ESSD300 model achieved such good results.

For future work we hope to apply this extension module to other computer vision tasks such as semantic segmentation. We are going to borrow some ideas from super-resolution to improve objection detection. We also plan to use more advanced technologies such as Generative Adversarial Networks to improve our detection results.

## ACKNOWLEDGMENT

The work in this paper is supported in part by the Natural Science Foundation of Shenzhen under Grant JCYJ20160506172651253, in part by the National Science and Technology Support Program under Grant 2015BAK01B04. We would like to thank the previous researchers for their outstanding achievements. Thanks for Professor Zhao's instruction and help.

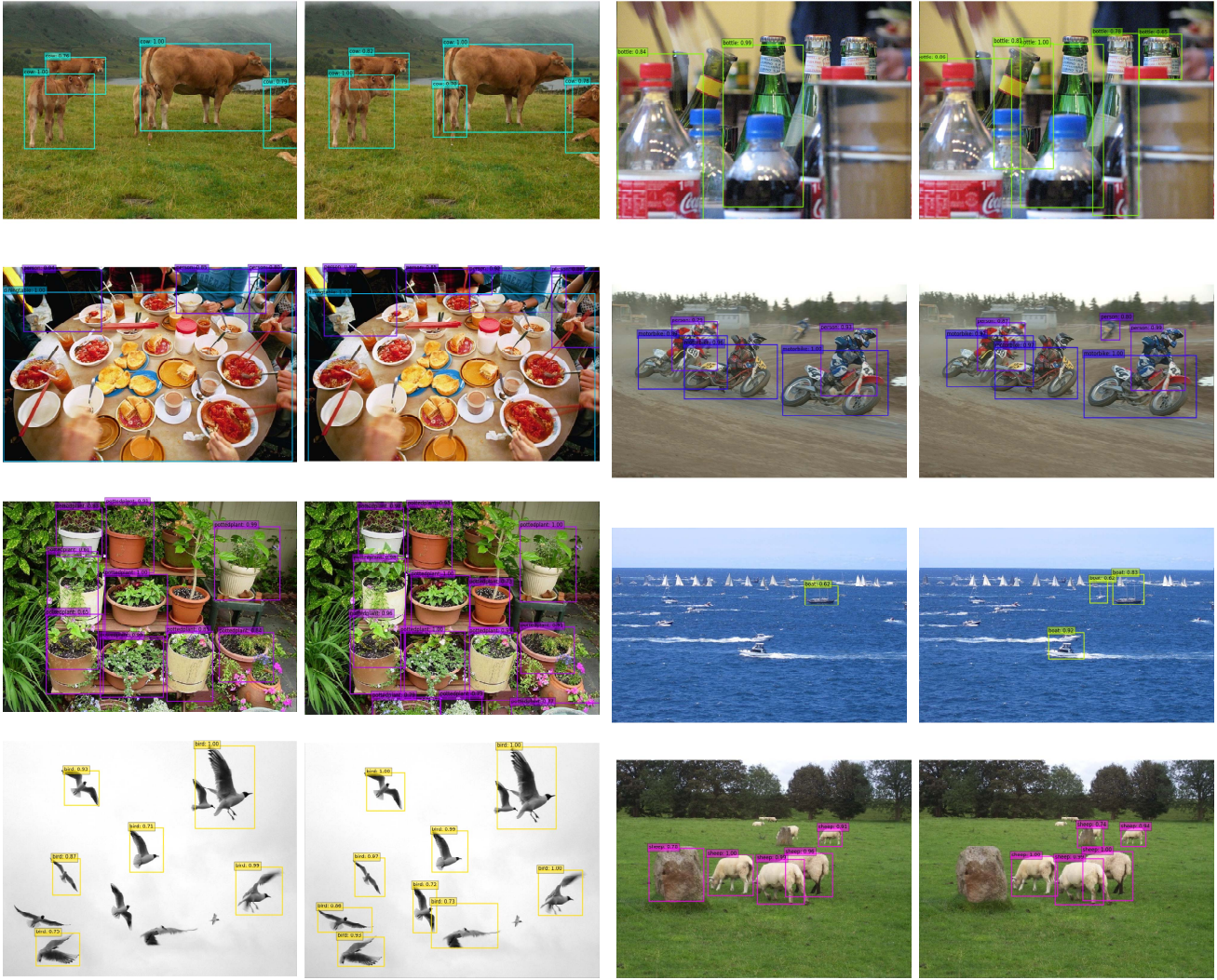

Figure 3. For each group, the left side is the result of SSD300 [7] and right side is the result of our ESSD.